%% file: main.tex
\documentclass[10pt,twocolumn,letterpaper]{article}

\usepackage{cvpr}              

\input{preamble}

\definecolor{cvprblue}{rgb}{0.21,0.49,0.74}
\usepackage[pagebackref,breaklinks,colorlinks,citecolor=cvprblue]{hyperref}
\usepackage{url}

\def\paperID{6054} 
\def\confName{CVPR}
\def\confYear{2025}

\title{MIMO: Controllable Character Video Synthesis with Spatial Decomposed Modeling}

\author{Yifang Men, Yuan Yao, Miaomiao Cui, Liefeng Bo \\
{\tt\small Tongyi Lab, Alibaba Group} \\
{\tt\small \url{https://menyifang.github.io/projects/MIMO/index.html}} \\
}

\begin{document}

\twocolumn[{%
\renewcommand\twocolumn[1][]{#1}%
\maketitle

\vspace{-20 pt}
\begin{center}
    \centering
  \setlength{\abovecaptionskip}{0.1cm}
 \setlength{\belowcaptionskip}{0.2cm}
    \includegraphics[width=1.0\linewidth]{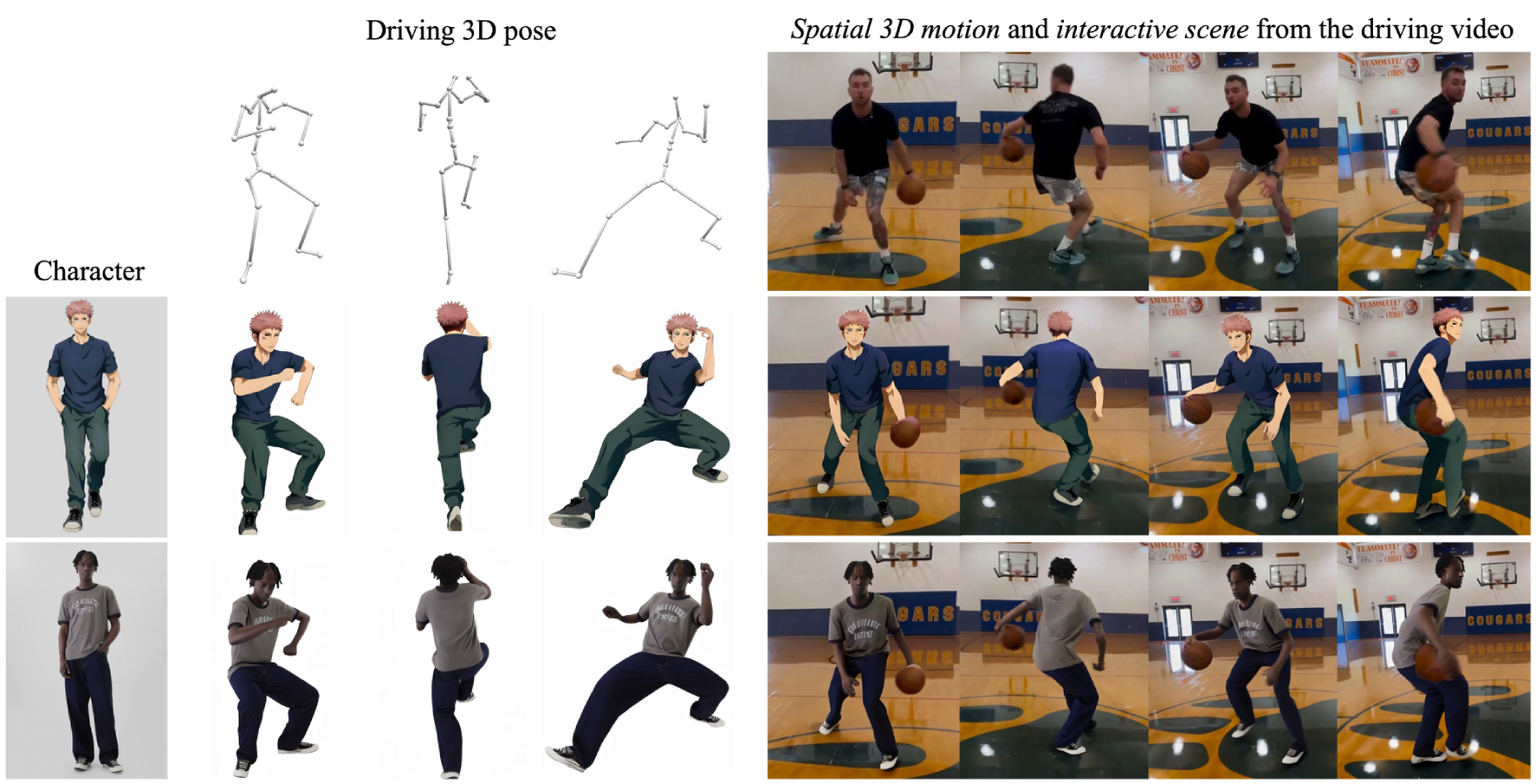}
    \captionof{figure}{Given a single reference image of character, MIMO can synthesize animated avatars in driving 3D poses (visualized as skeleton sequences) retrieved from motion datasets (left) or extracted from in-the-wild videos (right). Real-world scenes from driving videos can also be integrated into the synthesis with natural human-object interactions. MIMO simultaneously achieves advanced scalability to arbitrary characters, generality to novel 3D motions, and applicability to interactive real-world scenes in a unified framework. 
}
    \label{fig:teaser}
\end{center}%
}]

\begin{abstract}
Character video synthesis aims to produce realistic videos of animatable characters within lifelike scenes. As a fundamental problem in the computer vision and graphics community, 3D works typically require multi-view captures for per-case training, which severely limits their applicability of modeling arbitrary characters in a short time. Recent 2D methods break this limitation via pre-trained diffusion models, but they struggle for flexible controls, pose generality and scene interaction. To this end, we propose MIMO, a novel framework which can not only synthesize realistic character videos with controllable attributes (i.e., character, motion and scene) provided by simple user inputs, but also simultaneously achieve advanced scalability to arbitrary characters, generality to novel 3D motions, and applicability to interactive real-world scenes in a unified framework. The core idea is to encode the 2D video to compact spatial codes, considering the inherent 3D nature of video occurrence. Concretely, we lift the 2D frame pixels into 3D using monocular depth estimators, and decompose the video clip into three spatial components (i.e., main human, underlying scene, and floating occlusion) in hierarchical layers based on the 3D depth. These components are further encoded to canonical identity code, structured motion code and full scene code, which are utilized as control signals of the synthesis process. The design of spatial decomposed modeling enables flexible user control, complex motion expression, as well as 3D-aware synthesis for scene interactions. Experimental results show that the proposed method outperforms prior works by a large margin in character animation synthesis and is effective in providing a high degree of controllability (i.e., arbitrary characters, novel 3D motions, interactive scenes), thus enabling brand-new editing tasks (e.g., video character replacement). 

\end{abstract}

\section{Introduction}

Character video synthesis, an essential topic in areas of Computer Vision and Computer Graphics, has huge potential applications for movie production, virtual reality, and animation. While recent video generative models \cite{ho2022video,hong2022cogvideo,blattmann2023align, wu2023tune,zhang2023i2vgen,karras2023dreampose} have achieved great progress with text or image guidance, none of them fully captures the underlying attributes (e.g., appearance and motion of instance and scene) in a video and provides flexible user controls. Meanwhile, they still struggle for reasonable character synthesis in challenging scenarios, such as extreme 3D motions and complex object interactions accompanied by occlusions. 

\begin{figure}
\begin{center}
\setlength{\abovecaptionskip}{0.2cm}
\setlength{\belowcaptionskip}{-0.2cm}
\includegraphics[width=0.98\linewidth]{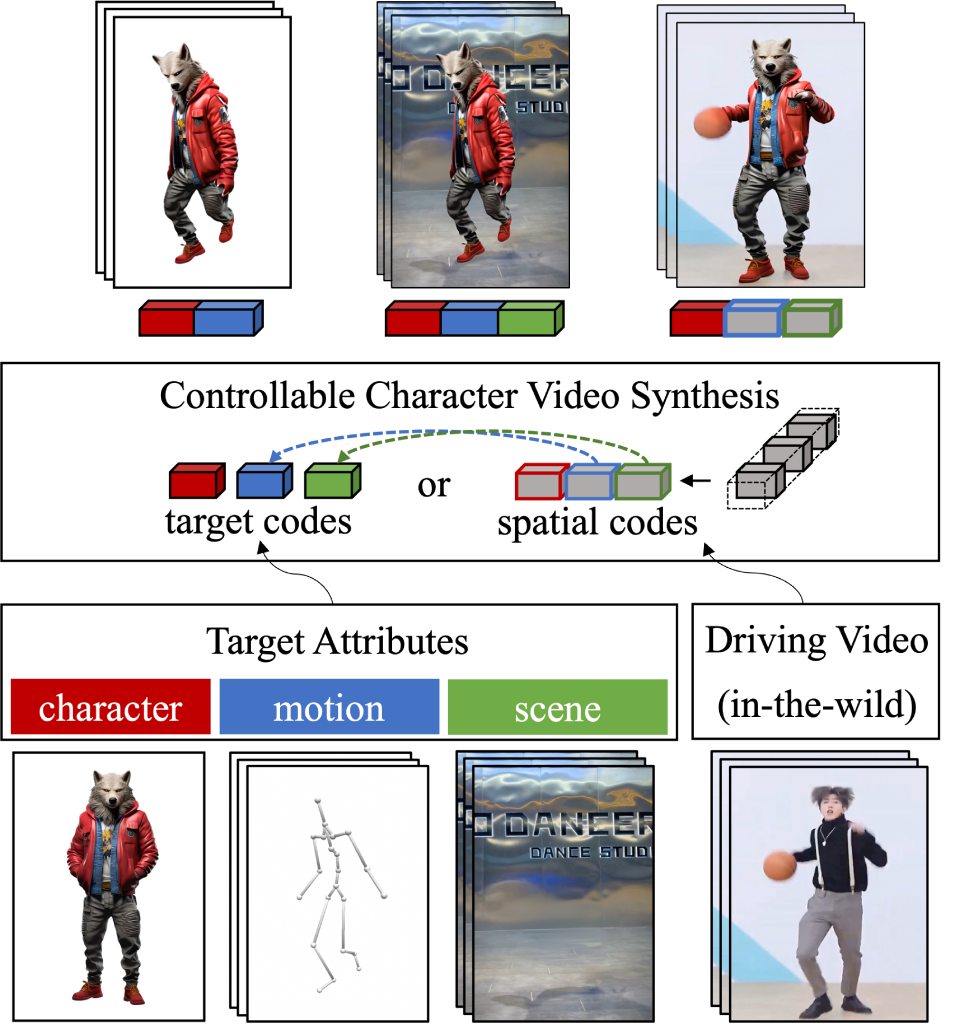}
\caption{
The basic idea of MIMO. Controllable character video synthesis with desired attributes provided by multiple inputs (e.g., a single image for character, a pose sequence for motion, and a single video even an image for scene) or a driving video. Target attributes are embedded into the latent space as the target codes and the driving video is spatially decomposed as the spatial codes. Target character videos can be generated in user control with the combined attribute codes. 
}
\label{fig:idea}
\end{center}
\end{figure}

The aim of this paper is to propose a brand-new and boosting method for controllable video synthesis, which can not only synthesize character videos with controllable attributes (i.e., character, motion and scene) provided by very simple user inputs, but also achieve advanced scalability to arbitrary characters, generality to novel 3D motions, and applicability to interactive real-world scenes in a unified framework (see Figure~\ref{fig:teaser}). 
In other words, the proposed method is capable of {\bf mi}micking anyone anywhere with complex {\bf m}otions and {\bf o}bject interactions, thus named {\bf MIMO}. 
As more concretely illustrated in Figure~\ref{fig:idea}, users are allowed to feed multiple inputs (e.g., a single image for character, a pose sequence for motion, and a single video even an image for scene) to provide desired attributes respectively or a direct driving video as input. The proposed model can embed target attributes into the latent space to construct target codes or encode the driving video with spatial-aware decomposition as spatial codes, thus enabling intuitive attribute control of the synthesis by freely integrating latent codes in a specific order.



Our task setting significantly decreases the cost of video creation and enables wide applications for not only character animation, but also video attribute editing (e.g., character replacement, motion transfer and scene insertion). 
However, it is extremely challenging due to the simplicity of user inputs, the complexity of real-world scenarios and the absence of 2D video annotations.
With the great progress of 3D neural representations (e.g., NeRF~\cite{mildenhall2021nerf} and 3D Gaussian splatting~\cite{kerbl20233d}), a series of works~\cite{peng2021animatable,liu2021neural,weng2022humannerf,hu2024gaussianavatar,li2024animatable} tend to represent the dynamic human as a pose-conditioned NeRF or Gaussian to learn animatable avatars in high-fidelity rendering quality.
However, they typically require fitting a neural field to multi-view captures or a monocular video of dynamic performers, which severely limits their applicability due to inefficient training and expensive data acquisition.
Another 3D works explored faster and cheaper solutions by directly inferring 3D models from single human images, following by rigged animation and physical rendering~\cite{huang2020arch,huang2024tech,liao2024tada,men2024en3d}. Unfortunately, the realism of the renderings is marginally compromised due to cumulative errors in sequential processes. 
Recently, several efforts~\cite{hu2024animate,xu2024magicanimate,zhu2024champ,wang2024disco} have investigated the potential of 2D diffusion models on image-guided character video synthesis, named character animation. They show that high-fidelity character video can be synthesized by inserting image feature extracted from the reference-net~\cite{hu2024animate,zhu2024champ} or control-net~\cite{xu2024magicanimate,zhang2023adding} into a pretrained diffusion model. 
However, they only focus on character synthesis in simple 2D motions (e.g., frontal dancing) and are less effective for articulated human motion in 3D space due to limited pose generality. Moreover, they fail to produce lifelike video for complicated scenes accompanied by human-object interactions and large camera movements.
We argue that the cause for these difficulties stems from insufficient video attribute parser considered only in 2D feature space, thereby disregarding the inherent 3D nature of video occurrence. 

To tackle these challenges, we propose a novel framework for controllable character video synthesis via spatial decomposed modeling. The core idea is to decompose and encode the 2D video in 3D-aware manner and employ more adequate expressions (e.g., 3D representations) for articulated properties. 
In contrast to previous works \cite{hu2024animate,zhang2024mimicmotion} directly learn the whole 2D feature of each video frame, we lift the 2D frame pixels into 3D, and construct the decomposed spatial representations in 3D space, which are equipped with richer contextual information and can be used for control signals of the synthesis process. 
Specifically, we decompose the video clip to three spatial components (scene, human and occlusion) in hierarchical layers based on 3D depth. In particular, human represents the main object in the video, scene represents the underlying background, and occlusion traces floating foreground objects.  For the human component, we further disentangle the identity property via canonical appearance transfer and encode the 3D motion representation via structural body codes. The scene and occlusion components are embedded with a shared VAE encoder and re-organized as a full scene code. 
The decomposed latent codes are inserted as conditions of a diffusion-based decoder to reconstruction the video clip. 
In this way, the network learns not only controllable synthesis of various attributes, but also 3D-aware layer composition of main object, foreground and background. 
Thereby, it enables flexible user controls as well as challenging cases of complicated 3D motions and natural object interactions.
In summary, our contributions are threefold:


\begin{itemize}
\item We propose a brand-new task that synthesizes character videos with controllable attributes by directly providing simple user inputs, and solve it with a novel approach to simultaneously achieve advanced scalability to arbitrary characters, generality to novel 3D motions, and applicability to interactive scenes in a unified framework. 
\item We introduce the spatial decomposed modeling, an effective architecture to simulate intricate video observations by encoding the inherent spatial components. It enables not only flexible user control, but also 3D-aware synthesis in human-object interaction contexts.
\item We tackle the challenge of inadequate pose representation for articulated human by introducing structured motion codes. 
It provides better expressive ability to handle complicated motions in spatial space, thus enabling advanced generality of generative model to novel 3D motions.
\end{itemize}

\begin{figure*}
\begin{center}
\setlength{\abovecaptionskip}{0cm}
\setlength{\belowcaptionskip}{-0.2cm}
\includegraphics[width=0.98\linewidth]{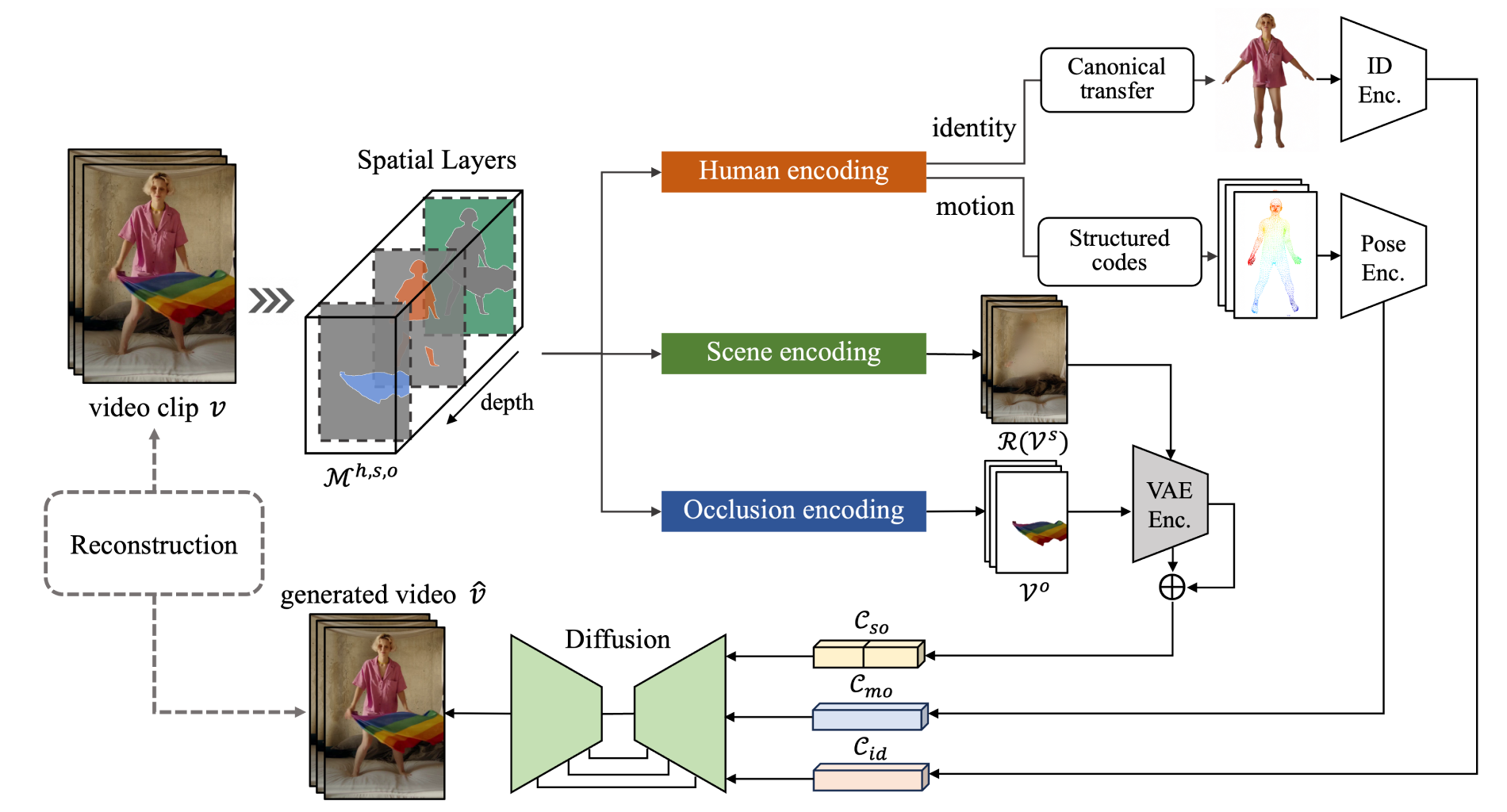}
\caption{
An overview of the proposed framework. The video clip is decomposed to three spatial components (i.e., main human, underlying scene, and floating occlusion) in hierarchical layers based on 3D depth. 
The human component is further disentangled for properties of identity and motion via canonical appearance transfer and structured body codes, and encoded to identity code $\mathcal{C}_{id}$ and motion code $\mathcal{C}_{mo}$. 
The scene and occlusion components are embedded with a shared VAE encoder and re-organized as a full scene code $\mathcal{C}_{so}$. These latent codes are inserted into a diffusion-based decoder as conditions for video reconstruction.
} 
\label{fig:network}
\end{center}
\end{figure*}


\section{Related Work}

\noindent \textbf{3D Human Modeling}.
Since the introduction of Neural Radiance Fields (NeRF)~\cite{mildenhall2021nerf}, neural human
representation has achieved remarkable success in obtaining articulated human models by fitting implicit neural fields to multi-view captures~\cite{peng2021neural,liu2021neural,su2021nerf} or a monocular video~\cite{weng2022humannerf,jiang2022neuman,jiang2022selfrecon}. 
HumanNeRF~\cite{weng2022humannerf} proposes to
represent human from a single video of a moving person by optimizing canonical volume and motion fields. NeuMan~\cite{jiang2022neuman}
jointly learns the decomposition of the human and the scene capable of novel pose rendering and animation of human in
the scene. HOSNERF~\cite{liu2023hosnerf} extend to support human-environment interactions by introduce a dynamic
human-object model. Recent works~\cite{li2024animatable,hu2024gaussianavatar} further introduce 3D Gaussian splatting~\cite{kerbl20233d} for realistic renderings. 
Despite achieving promising results, these methods typically require expensive data acquisition with precise camera/pose estimates and are less efficient for training and rendering, severely limiting their real-world applications.
Another series of work explored faster and cheaper human modeling solutions via generative models ~\cite{hong2022eva3d,yang20233dhumangan,dong2023ag3d,men2024en3d}. 
En3D~\cite{men2024en3d} learns an enhanced 3D human generator with efficient refiner to directly infer 3D models from single images or text prompts in few minutes. 
These generated avatars are rendered in a canonical body pose and aligned to an underlying 3D skeleton which allows for easy animation and the generation of motion videos. However, the fidelity of the driven results
is marginally compromised due to cumulative errors inherent in the rendering processes.

\noindent \textbf{Diffusion-based Character Video Synthesis}.
The remarkable progress in diffusion models has demonstrated promising results in image and video generation~\cite{ho2020denoising,rombach2022high,ho2022video,hong2022cogvideo,blattmann2023align}. 
A plethora of methodologies proposed to incorporate these pre-trained diffusion models for human-centric video synthesis~\cite{wang2023disco,ma2024follow,feng2023dreamoving,hu2024animate,xu2024magicanimate,zhu2024champ,zhang2024mimicmotion},
enabling the transformation of character images into animated videos controlled by desired pose sequences.
MagicAnimate~\cite{xu2024magicanimate} utilizes ControlNet~\cite{zhang2023adding} and an appearance encoder for identity preservation and pose guidance,
building upon a video diffusion model. Animate Anyone~\cite{hu2024animate} employs a UNet-based ReferenceNet to extract detailed features
from reference images. Champ~\cite{zhu2024champ} further introduces a 3D parametric model to extract motion guidance as conditions. MimicMotion~\cite{zhang2024mimicmotion} presents a confidence-aware pose guidance approach to enhance
generation quality and temporal smoothness. Despite producing visually appealing results, these methods encounter quality degradation issues in complex motion scenarios and are incapable of handling occlusion-aware generation in human-object interaction contexts. 
Our method built on diffusion models overcomes these challenges by a novel generative architecture with spatial decomposed modeling, considering inherent 3D nature for 2D videos.


\section{Method Description}


Our goal is to synthesize high-quality character videos with user-controlled visual attributes, such as character, motion and scenes. The desired attributes can be automatically extracted from an in-the-wild character video or simply provided by a single image, a pose sequence, and a single video, respectively. Different from previous methods using only weak control signals (e.g., text prompt) \cite{zhao2023make,ma2024follow} or insufficient 2D expressions \cite{hu2024animate,zhang2024mimicmotion}, our model achieves automatic and unsupervised separation of spatial components and encodes them into compact latent codes considering inherent 3D nature to control the synthesis. 
Thus, our dataset can only contain 2D character videos $\{v \in \mathbb{R}^{N \times H \times W}\}$ without any annotations. 

The overview of the proposed framework is illustrated in Figure~\ref{fig:network}. Given a video clip $v$, MIMO learns a reconstruction process with automatic attribute encoding and composed condition decoding. Considering 3D nature of video occurrence, we extract the three spatial components in hierarchical layers based on 3D depth (Section~\ref{sec:decomp}). The first component of human is encoded with disentangled properties of identity and motion (Section~\ref{sec:enc_human}). The last two components of scene and occlusion are embedded with a shared encoder and re-organized as a scene code (Section~\ref{sec:enc_scene}). These latent codes $\mathcal{C}$ are inserted into a diffusion decoder $\mathcal{D}$ as composed conditions (Section~\ref{sec:dec}). $\mathcal{C}$, $\mathcal{D}$ are jointly learned by minimizing the difference between the generated frames and input frames via noise prediction (Section~\ref{sec:training}). 

\subsection{Hierarchically spatial layer decomposition}
\label{sec:decomp}

Considering the inherent 3D elements of video composition, we split a video $v=\{\mathcal{I}_t | t=1,…,N\}$ into three main components: human as a core performer, scene as the underlying background, and occluded object as the floating foreground. To automatically decompose them, we lift 2D pixels into 3D and track detected objects in hierarchical layers based on corresponding depth values. 

To start with, for each frame $\mathcal{I}_t \in v$, we obtain its monocular depth map using a pretrained monocular depth estimator \cite{yang2024depth}. The human layer is firstly extracted with human detection \cite{wu2019detectron2}, and propagate to video volume via video tracking method \cite{ravi2024sam}, thus obtaining $\mathcal{M}^h \in R^{N*H*W}$, a binary mask sequence along the time axis (i.e., masklet). Subsequently, we extract the occlusion layer with objects whose mean depth values are smaller than the human layer, and generate masklet predictions $\mathcal{M}^o$ via a video tracker. The scene layer can be obtained by removing human and occlusion objects, defined by scene masklet $\mathcal{M}^s$. With predicted masklets, we can compute the decomposed human video of component $i$ by multiplying the original source video with component masklet $\mathcal{M}^i$:
\begin{equation}
{v}^i= {v} \odot \mathcal{M}^i, i=\{h,o,s\},
\end{equation}
where $\odot$ denotes element-wise product. $v^i$ is then fed into the corresponding branch for human, scene and occlusion encoding, respectively.

\subsection{Disentangled human encoding}
\label{sec:enc_human}

This branch aims to encode the human component $v^h$ into the latent space as disentangled codes $\mathcal{C}_{id}$ and $\mathcal{C}_{mo}$ of identity and motion. Previous works \cite{hu2024animate,xu2024magicanimate,zhang2024mimicmotion} typically random select one frame from the video clip as appearance representation, and employ extracted 2D skeleton with key-points as the pose representation. Essentially, this design exists two core issues which may limit networks’ performance: 1) It is hard for 2D pose to adequately express motions which take place in 3D spatial space, especially for articulated ones accompanied by exaggerated deformations and frequent self-occlusions. 2) The postures of frames across a video are highly similar, and there inevitably exists the entanglement between appearance frame and target frame both retrieved from the posed video. Thereby, we introduce new 3D representations of motion and identity for adequate expression and full disentanglement. 

\noindent {\bf Structured motion code.}
We define a set of latent codes $\mathcal{Z}=\{z_1,z_2,…,z_{6890}\}$, and anchor them to corresponding vertices of a deformable human body model (SMPL) \cite{SMPL2015}.
For the frame $t$, SMPL parameters $\mathcal{S}_t$ and camera parameters $\mathcal{C}_t$ are estimated from the monocular video frame $v^h_t$ using \cite{goel2023humans}. The spatial locations of the latent codes are then transformed based on the human pose $\mathcal{S}_t$ and projected to the 2D plane based on the camera setting $\mathcal{C}_t$. Using a differentiable rasterizer \cite{Laine2020diffrast} with vertex interpolation, the 2D feature map $\mathcal{F}_t$ in continuous values can be obtained. $\{\mathcal{F}_t, t=1,..,N\}$ will be stacked along the time axis and embedded into the latent space as the motion code $\mathcal{C}_{mo}$ by a pose encoder $\mathcal{E}_p$. In this way, we establish correspondences between the same set of identifiable latent codes on underlying 3D body surface and posed 2D renderings at different frames of arbitrary videos. 
This structured motion code enables clearer and more dense pose representation for articulated 3D motions in spatial space.

\noindent {\bf Canonical appearance transfer.}
To fully disentangle the appearance from posed video frames, an ideal solution is to learn the dynamic human representation from the monocular video and transform it from the posed space to the canonical space. Considering the efficiency, we employ a simplified method that directly transforms the posed human image to the canonical result in standard A-pose using a pretrained human repose model. The synthesized canonical appearance image is fed to ID encoders to obtain the identity code $\mathcal{C}_{id}$. This simple design enables full disentanglement of identity and motion attributes. Following \cite{hu2024animate}, the ID encoders include a CLIP image encoder and a reference-net architecture to embed for the global and local feature, respectively, which composes $\mathcal{C}_{id}$.

\subsection{Scene and occlusion encoding}
\label{sec:enc_scene}

In scene and occlusion branches, we use a shared and fixed VAE encoder \cite{kingma2013auto} to embed the $v^s$ and $v^o$ into the latent space as the scene code $\mathcal{C}_s$ and occlusion code $\mathcal{C}_o$, respectively. Before $v^s$ input, we pre-recover it by a video inpainting method \cite{zhou2023propainter} as $\mathcal{R}(v^s)$ to avoid the interference brought by mask contours. Then the scene code $\mathcal{C}_s$ and the occlusion code $\mathcal{C}_o$ are concatenated together in order to get the full scene code $\mathcal{C}_{so}$ for composed synthesis. The independent encoding of spatial components (i.e., middle human, underlying scene, and floating occlusion) enable the network to learn an automatic layer composition, thus achieving natural character insertion in complicated scenes even with occluded object interactions.

\begin{figure}
\begin{center}
\setlength{\abovecaptionskip}{-0.1cm}
\setlength{\belowcaptionskip}{-0.6cm}
\includegraphics[width=0.98\linewidth]{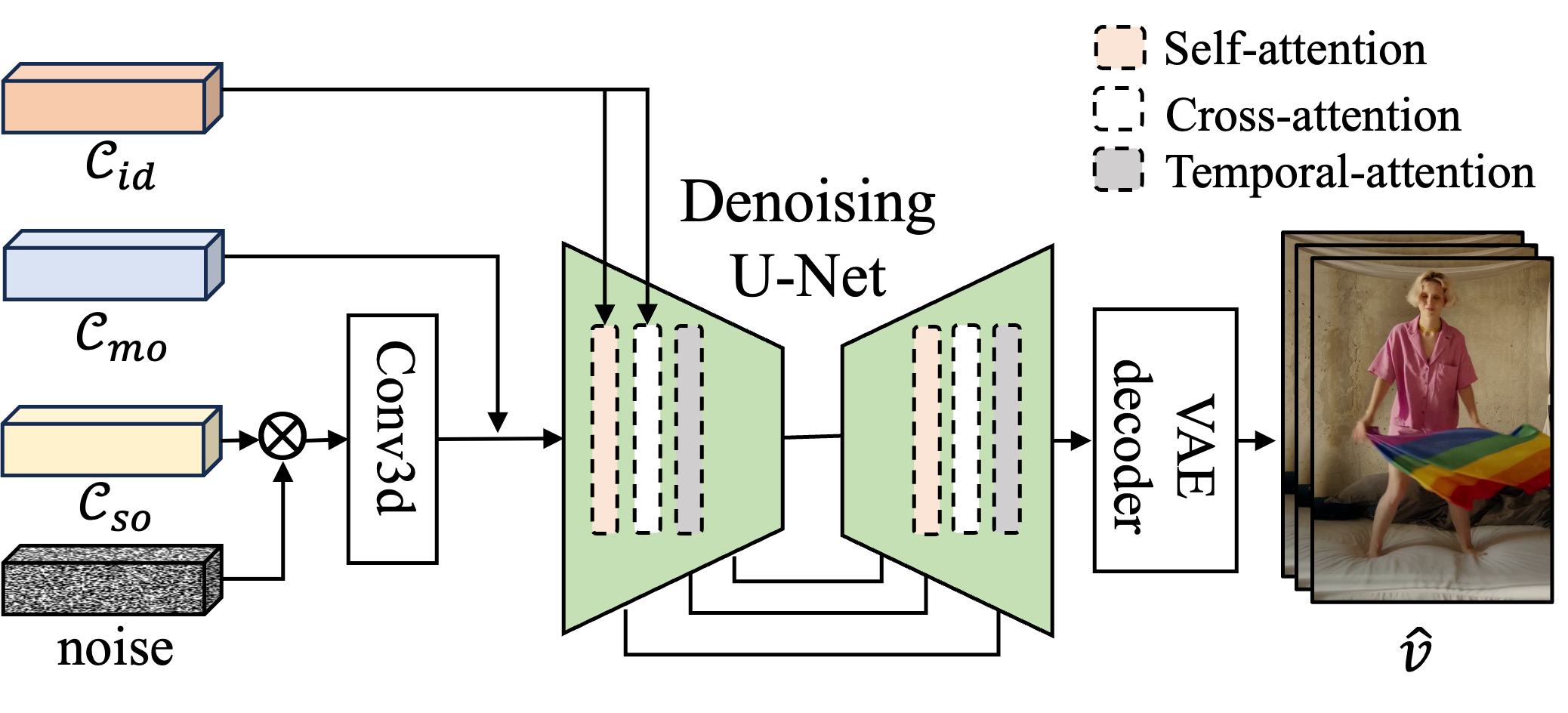}
\caption{The architecture of the diffusion-based decoder.} 
\label{fig:decoder}
\end{center}
\end{figure}

\begin{figure*}
\begin{center}
\setlength{\abovecaptionskip}{0cm}
\setlength{\belowcaptionskip}{-0.6cm}
\includegraphics[width=1\linewidth]{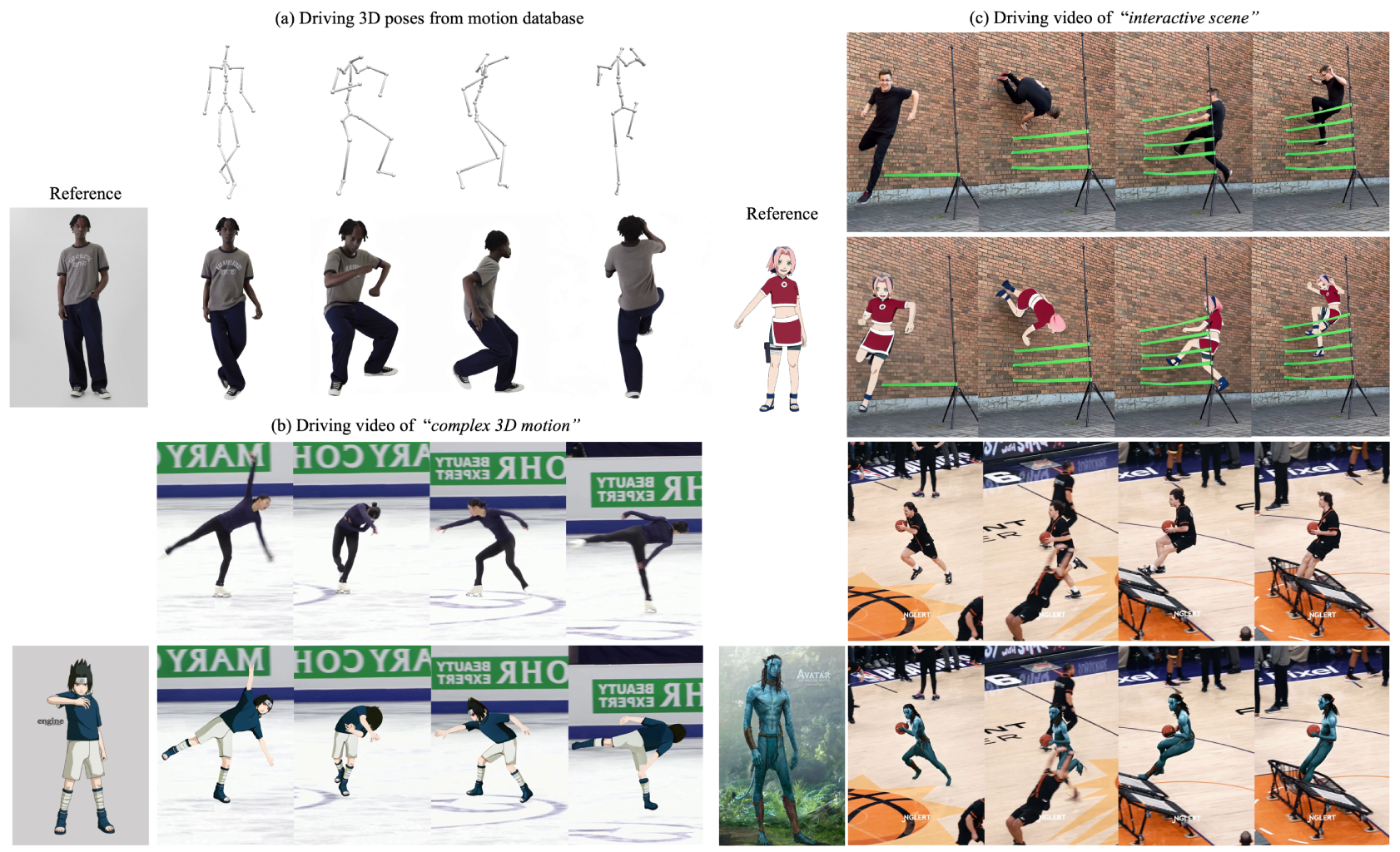}
\caption{
Results of animating diverse characters (e.g., realistic humans, cartoon characters and personified ones) with novel 3D motions retrieved from the motion database (a) or extracted from the driving video (b), and interactive scenes from in-the-wild videos (c). 
} 
\label{fig:res_ability}
\end{center}
\end{figure*}

\subsection{Composed decoding}
\label{sec:dec}

Given the latent codes of decomposed attributes, we re-compose them as conditions of the diffusion-based decoder for video reconstruction. As shown in Figure~\ref{fig:decoder}, we adapt denoising U-Net backbone built upon Stable Diffusion (SD) \cite{rombach2021highresolution} with temporal layers from \cite{guo2023animatediff}. The full scene code $\mathcal{C}_{so}$ is concatenated with the latent noise, and is fed into a 3D convolution layer for fusion and alignment. The motion code $\mathcal{C}_{mo}$ is added to the fused feature and input to the denoising U-Net. For identity code $\mathcal{C}_{id}$, its local feature and global feature are inserted into the U-Net via self-attention layers and cross-attention layers as~\cite{hu2024animate}, respectively. Finally, the denoised result is converted into the video clip $\hat{v}$ via a pretrained VAE decoder \cite{kingma2013auto}.

\subsection{Training}
\label{sec:training}

For the training, we employ the diffusion noise-prediction loss to simulate video reconstruction process:
\begin{equation}
\mathcal{L}= \mathbb{E}_{x_0,c_{id},c_{so},c_{mo},t,\epsilon \in \mathcal{N}(0,1)}[||\epsilon-\epsilon_\theta(x_t, c_{id},c_{so},c_{mo},t)||^2_2]
\end{equation}
where $x_0$ is the augmented input sample, $t$ denotes the diffusion timestep, $x_t$ is the noised sample at $t$, and $\epsilon_\theta$ represents the function of the denoising UNet. 

\noindent {\bf Implementation details.}
Our method is implemented in PyTorch using 8 NVIDIA Tesla-A100 GPUs with 80GB memory. We initialize the model of denoising U-Net and reference-net based on the pre-trained weights from SD 1.5 \cite{rombach2021highresolution}, whereas the motion module is initialized with the weights of AnimateDiff \cite{guo2023animatediff}. During training, the weights of VAE encoder and decoder, as well as the CLIP image encoder are frozen. We optimize the denoising U-Net, pose encoder and reference-net with the diffusion noise-prediction loss. It takes around 50k iterations with 24 video frames and a batch size of 8 for converge. 


\begin{figure*}
\begin{center}
\setlength{\abovecaptionskip}{-0.1cm}
\setlength{\belowcaptionskip}{-0.6cm}
\includegraphics[width=1\linewidth]{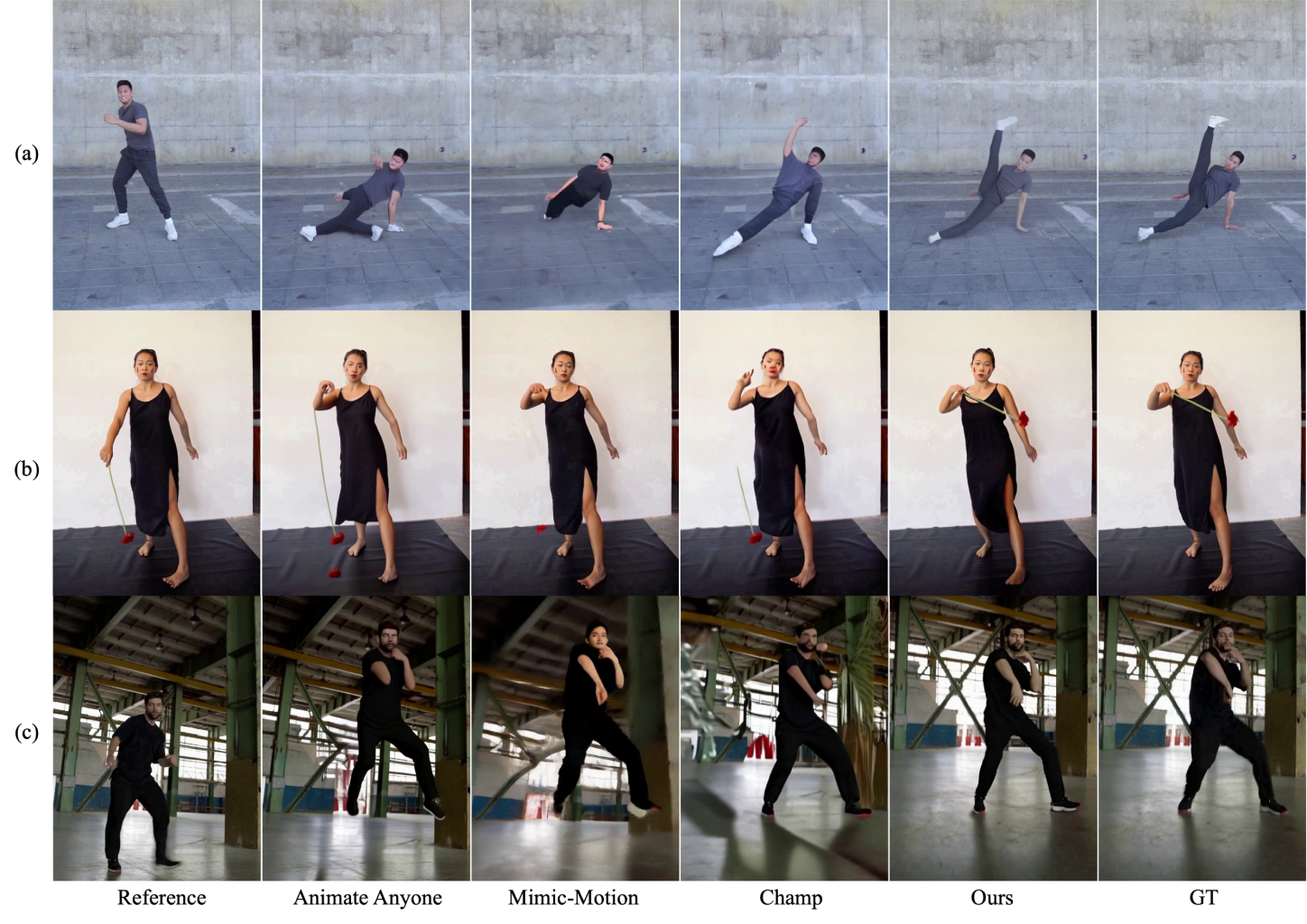}
\caption{Qualitative comparison with three state-of-the-art methods: Animate Anyone~\cite{hu2024animate}, Mimic-Motion~\cite{zhang2024mimicmotion} and Champ~\cite{zhu2024champ}.} 
\label{fig:res_compare}
\end{center}
\end{figure*}

\section{Experimental Results}

\noindent {\bf Dataset.} 
We create a human video dataset called HUD-7K to train our model. This dataset consists of $5K$ real character videos and $2K$ synthetic character animations. The former does not require any annotations and can be automatically decomposed to various spatial attributes via our scheme. To enlarge the range of the real dataset, we also synthesize $2K$ videos by rendering character animations in complex motions under multiple camera views, utilizing En3D \cite{men2024en3d}. These synthetic videos are equipped with accurate annotations due to completely controlled production. 
For the evaluation, we collect 100 in-the-wild human videos covering diverse contents (e.g., dancing, sports and movie) and randomly truncate them to 150-frame clips as test set.

 \noindent {\bf Metrics.} 
We follow~\cite{hu2024animate} to evaluate our method using four standard metrics: Peak Signal-to-Noise Ratio (PSNR)~\cite{hore2010image}, Structural Similarity Index Measure (SSIM)~\cite{wang2004image}, Learned Perceptual Image Patch Similarity (LPIPS)~\cite{zhang2018unreasonable} for image-level quality, and Fr\'echet Video Distance (FVD)~\cite{unterthiner2018towards} for video-level evaluation.

\subsection{Controllable character video synthesis}

Given the target attributes of character, motion and scene, our method can generate realistic video results with their latent codes combined for guided synthesis. The target attributes can be provided by simple user inputs (e.g., single images/videos for character/scene, pose sequences from large database \cite{mahmood2019amass,mixamo2018} for motion) or flexibly extracted from the real-world videos, involving complicated scenes of occluded object interactions and extreme articulated motions. In the following, MIMO demonstrates that it can simultaneously achieve advanced scalability to arbitrary characters, generality to novel 3D motions, and applicability to in-the-wild scenes in a unified framework. More results can be found in the supplemental materials (Supp).

\noindent {\bf Arbitrary character control.}
As shown in Figure~\ref{fig:res_ability}, our method can animate arbitrary characters, including realistic humans, cartoon characters and personified ones. Various body shapes of characters can be faithfully preserved due to the decoupled pose and shape parameters in structured motion representation.

\noindent {\bf Novel 3D motion control.}
To verify the generality to novel 3D motions, we test MIMO using challenging out-of-distribution pose sequences from the AMASS~\cite{mahmood2019amass} and Mixamo~\cite{mixamo2018} database, including dancing, playing and climbing (Figure~\ref{fig:res_ability} (a)). We also try complex spatial motions by extracting them from in-the-wild videos for driving (Figure ~\ref{fig:res_ability} (b, c)). Our method exhibits high robustness for these novel 3D motions under different viewpoints. 

\noindent {\bf Interactive scene control.}
We validate the applicability of our model to complicated real scenes by extracting both scene and motion attributes from in-the-wild videos for character animation, as a brand-new task of video character replacement. Figure~\ref{fig:res_ability} (c) shows that the characters can be seamlessly inserted to the real scenes with natural human-object interactions.

\subsection{Comparison with state-of-the-arts}

\noindent {\bf Qualitative comparison.}
In Figure~\ref{fig:res_compare}, we compare the synthesis results of our method with three state-of-the-art character animation methods: Animate Anyone~\cite{hu2024animate}, Mimic-Motion~\cite{zhang2024mimicmotion} and Champ~\cite{zhu2024champ}. All the results of these methods are obtained by using the source codes and trained models released by authors or popular re-implements, following by fine-tuning in our training dataset. As we can see, all previous methods fail to produce extreme articulated human motions with exaggerated deformations and frequent self-occlusions (Figure~\ref{fig:res_compare} (a)). They also 
cannot handle complicated scenes of object interaction (Figure~\ref{fig:res_compare} (b)) and large camera movement (Figure~\ref{fig:res_compare} (c)). In contrast, our method tackles these challenges and gives more realistic results in both global structures and detailed textures. More video results can be found in Supp. Furthermore, our method shows its superiority that it enables directly inferring animatable avatars in free-viewpoint with inter-frame consistency to some extent, which are comparable to the results of SOTA training-based 3D method, as presented in Supp.

\noindent {\bf Quantitative comparison.}
Table~\ref{table:quantitative} shows the comparison of our method with~\cite{hu2024animate, zhang2024mimicmotion, zhu2024champ} in terms of the PSNR, SSIM, LPIPS and FVD metrics, respectively. Due to the presence of a certain quantity of complex cases (e.g., including spatial motions, scene interactions, and camera movements) in the test set, it’s extremely challenging to model the intricate interplay of real-world scenarios. 
Even so, our method demonstrated the best performance in these metrics among all methods. It outperforms previous works by a margin of at least 4.16 in terms of PSNR and 0.152 in terms of SSIM, etc. In contrast to directly learning the entire 2D video frame with only inadequate human pose annotations, MIMO decompose 2D frames into hierarchically spatial components with more expressive 3D representations. The results indicate that our method better simulates video observations of the real physical world.
Considering insufficient scene modeling of previous methods, we also provide additional quantitative comparison by removing background and object for only character synthesis in Supp.

\begin{table}
\setlength{\abovecaptionskip}{0.1cm}
  \centering
    \caption{Quantitative comparison with state-of-the-art methods in terms of PSNR, SSIM, LPIPS and FVD. 
    }
    \small
  \begin{tabular}{ccccc}
    \toprule
    Method &  PSNR$\uparrow$  &  SSIM$\uparrow$  & LPIPS$\downarrow$ & FVD$\downarrow$  \\
    \midrule
    Animate Anyone~\cite{hu2024animate} & 21.003 & 0.722 & 0.264  & 304.3 \\
    Mimic-Motion~\cite{zhang2024mimicmotion} & 20.688 & 0.731  & 0.343  & 289.2 \\
    Champ~\cite{zhu2024champ} & 21.044 & 0.724 & 0.312  & 412.5 \\
    Ours & \textbf{25.210}  & \textbf{0.883} & \textbf{0.125} & \textbf{221.4}  \\
    \bottomrule
  \end{tabular}
  \label{table:quantitative}
\end{table}

\begin{figure}
\begin{center}
\setlength{\abovecaptionskip}{-0.1cm}
\setlength{\belowcaptionskip}{-0.6cm}
\includegraphics[width=0.98\linewidth]{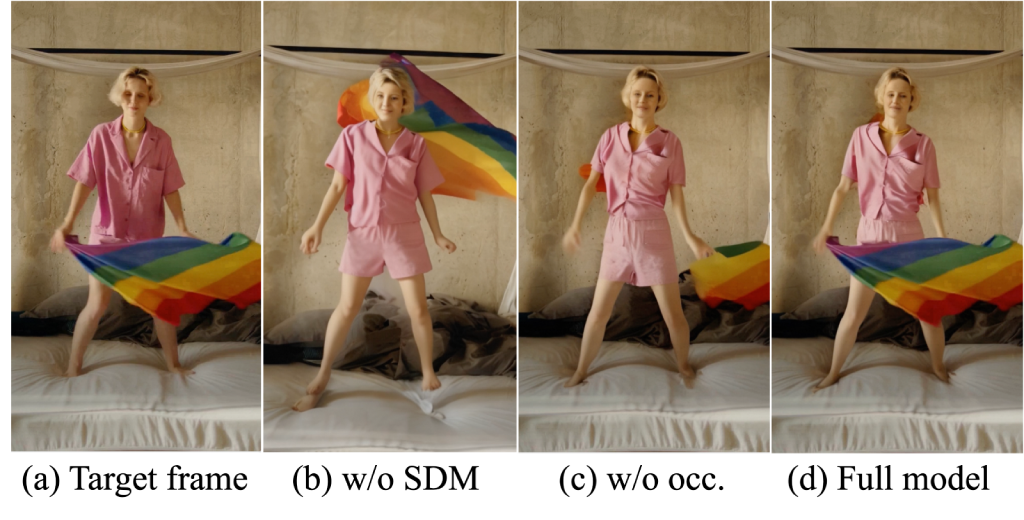}
\caption{Effects of spatial decomposed modeling.} 
\label{fig:res_ab_sdm}
\end{center}
\end{figure}

\subsection{Ablation study}

\noindent {\bf Spatial decomposed modeling.} We assess the impact of this design by training a model via randomly selecting one frame from videos as the appearance reference without decomposed layers (w/o SDM). In this way, it fails to produce faithful background and interactive foreground, easily suffering from unstable texture distortions for large camera movements (Figure~\ref{fig:res_ab_sdm} (a, b)). Essentially, this instability stems from the absent guidance of scene generation, relying only on weak correlation between the scene and character movement revealed by data distribution. We also attempt to model the human and mixed scene without independent occlusion encoding (w/o occ.), and Figure~\ref{fig:res_ab_sdm} (c, d) shows that this variation cannot synthesize reasonable occluded objects for scene interaction without the ability to comprehend spatial layers. Figure~\ref{fig:res_ab_sdm} (d) and Table~\ref{table:ablation} also indicates that this decomposed strategy yield more realistic results with facial details and clothing wrinkles.

\begin{table}
\setlength{\abovecaptionskip}{0.1cm}
  \centering
    \caption{
    Results of models trained by removing specific modules or replacing with alternative designs for ablation study.
    }
    \small
  \begin{tabular}{ccccc}
    \toprule
    Method &  PSNR$\uparrow$  &  SSIM$\uparrow$  & LPIPS$\downarrow$ & FVD$\downarrow$  \\
    \midrule
    Ours-w/o SDM & 22.148 & 0.762 & 0.231  & 268.5 \\
    Ours-w/ 2D skeleton & 24.326 & 0.842 & 0.186  & 237.2 \\
    Ours-w/ 3D maps & 24.402 & 0.844 & 0.203  & 278.1 \\
    Ours-w/o CA & 24.918 & 0.871 & 0.148  & 223.1 \\
    Ours & \textbf{25.210}  & \textbf{0.883} & \textbf{0.125} & \textbf{221.4}  \\
    \bottomrule
  \end{tabular}
  \label{table:ablation}
\end{table}
  
\begin{figure}
\begin{center}
\setlength{\abovecaptionskip}{-0cm}
\setlength{\belowcaptionskip}{-0.7cm}
\includegraphics[width=0.98\linewidth]{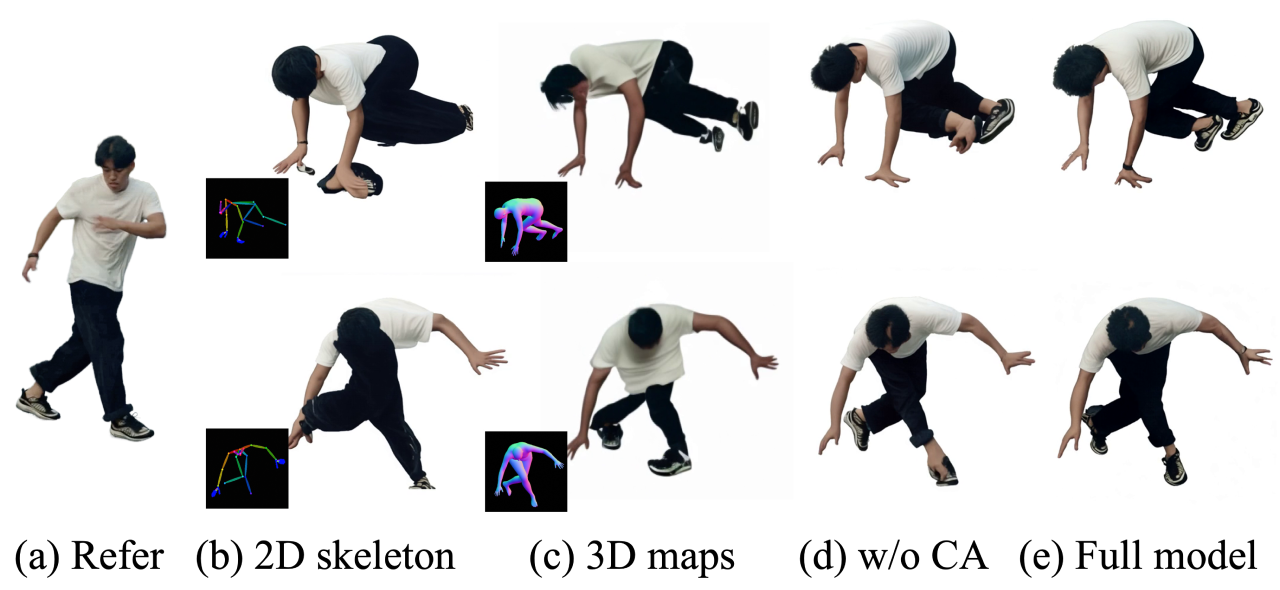}
\caption{Effects of structured motion representation and canonical appearance transfer.} 
\label{fig:res_ab_motion}
\end{center}
\end{figure}

\noindent {\bf Structured motion code.} To verify the effectiveness of the proposed structured motion representation (SMR), we evaluate the performance of several variants of our method by employing alternative motion formats: commonly used 2D skeleton in~\cite{hu2024animate} and 3D maps in~\cite{zhu2024champ}. As shown in Figure~\ref{fig:res_ab_motion} and Table~\ref{table:ablation}, 2D skeleton ignores the occlusion relationship in bones and muscles, resulting in ambiguity for spatial motions. The 3D maps, consisting of normal map, depth map, etc., improve the pose representation capability, but still struggle for highly complex spatial motions due to undefined labels of dense body parts.
Our SMR, inserting identifiable codes into structured body surfaces and projecting for 2D correspondence, 
provides strong articulation ability of motion in spatial space and significantly improves the model's generalizability to novel 3D motions. 

\noindent {\bf Canonical appearance transfer.}  This design (CA) further disentangles motion and identity in consecutive video frames with high posture correlation. It leads to more effective learning of SMR and 
obviously alleviates the issue of synthesis confusion between hands and feet (Figure~\ref{fig:res_ab_motion} (d)).

\section{Conclusions}
In this paper, we presented MIMO, a novel framework for controllable character video synthesis, which allows for flexible user control with simple attribute inputs. Our method introduces a new generative architecture which decomposes the 2D video to various spatial components, and embeds their latent codes as the condition of decoder to reconstruct the video. Experimental results demonstrated that our method enables not only flexible character, motion and scene control, but also advanced scalability to arbitrary characters, generality to novel 3D motions, and applicability to interactive scenes. We also believed that our solution, which considers inherent 3D nature of video occurrence and automatically encodes the 2D video to hierarchical spatial components could inspire future researches for 3D-aware video modeling. Furthermore, our framework is not only well suited to generate character videos but also can be potentially adapted to common video synthesis tasks.

{
    \small
    \bibliographystyle{ieeenat_fullname}
    \bibliography{main}
}


\end{document}

%% file: preamble.tex
\usepackage[dvipsnames]{xcolor}
